\documentclass{article}

\usepackage{microtype}
\usepackage{graphicx}
\usepackage{xcolor}
\usepackage{subfigure}
\usepackage{booktabs} 

\usepackage{hyperref}

\usepackage[accepted]{icml2024}

\usepackage{amsmath}
\usepackage{amssymb}
\usepackage{mathtools}
\usepackage{amsthm}

\DeclareMathOperator{\Embed}{Embed}
\newcommand{\catname}[1]{{\normalfont\textbf{#1}}}

\usepackage[capitalize,noabbrev]{cleveref}

\theoremstyle{plain}
\newtheorem{theorem}{Theorem}[section]

\theoremstyle{definition}
\newtheorem{definition}[theorem]{Definition}

\theoremstyle{remark}

\usepackage[textsize=tiny]{todonotes}

\usepackage{multirow}

\icmltitlerunning{Metric Learning for Clifford Group Equivariant Neural Networks}

\begin{document}

\twocolumn[
\icmltitle{Metric Learning for Clifford Group Equivariant Neural Networks}



\icmlsetsymbol{equal}{*}

\begin{icmlauthorlist}
\icmlauthor{Riccardo Ali}{equal,cambridge}
\icmlauthor{Paulina Kulytė}{equal,cambridge}
\icmlauthor{Haitz Sáez de Ocáriz Borde}{oxford}
\icmlauthor{Pietro Liò}{cambridge}
\end{icmlauthorlist}

\icmlaffiliation{cambridge}{University of Cambridge}
\icmlaffiliation{oxford}{University of Oxford}

\icmlcorrespondingauthor{Riccardo Ali}{rma55@cam.ac.uk}
\icmlcorrespondingauthor{Paulina Kulytė}{pk556@cam.ac.uk}

\icmlkeywords{metric learning, Clifford algebra, category theory}

\vskip 0.3in
]



\printAffiliationsAndNotice{\icmlEqualContribution} 

\begin{abstract}
Clifford Group Equivariant Neural Networks~(CGENNs) leverage Clifford algebras and multivectors as an alternative approach to incorporating group equivariance to ensure symmetry constraints in neural representations. In principle, this formulation generalizes to orthogonal groups and preserves equivariance regardless of the metric signature. However, previous works have restricted internal network representations to Euclidean or Minkowski (pseudo-)metrics, handpicked depending on the problem at hand. In this work, we propose an alternative method that enables the metric to be learned in a data-driven fashion, allowing the CGENN network to learn more flexible representations. Specifically, we populate metric matrices fully, ensuring they are symmetric by construction, and leverage eigenvalue decomposition to integrate this additional learnable component into the original CGENN formulation in a principled manner. Additionally, we motivate our method using insights from category theory, which enables us to explain Clifford algebras as a categorical construction and guarantee the mathematical soundness of our approach. We validate our method in various tasks and showcase the advantages of learning more flexible latent metric representations. The code and data are available at \href{https://github.com/rick-ali/Metric-Learning-for-CGENNs}{https://github.com/rick-ali/Metric-Learning-for-CGENNs}

\end{abstract}

\section{Introduction}

\textit{Clifford (or geometric) algebras} play a significant role in physics~\citep{Baylis2004ApplicationsOC}, where they have been used to represent operators such as spinors and Dirac matrices in quantum mechanics, Maxwell's equations in electromagnetism, and Lorentz transformations in relativity. They leverage addition and the geometric product as their two fundamental operations, providing a powerful mathematical language for expressing geometric concepts in a unified manner. In particular, they can describe rotations, reflections, translations, and other geometric transformations succinctly, which frequently arise in physical phenomena.

Recently, Clifford algebras have been introduced to deep learning: Clifford Group Equivariant Neural Networks~(CGENNs)~\citep{ruhe2023clifford} harness the mathematical framework of Clifford algebras to represent data in a way that maintains geometric symmetries and equivariance with respect to several groups such as $O(n)$, $SO(n)$, $E(n)$, and $SE(n)$. However, they currently support only diagonal and fixed metrics to model internal network representations. These typically involve the standard metric on Euclidean space, represented as \(Q_{E}=\text{diag}(1,\dots,1)\), and the Minkowski pseudo-metric, \(Q_{M}=\text{diag}(-1,\dots,1)\), which, in principle, have inherent physical significance but must be chosen a priori. Ideally, we would like to enable the model to learn as rich internal representations as possible without being constrained to diagonal metric matrices. Thus, inspired by recent work on latent trainable geometries~\cite{de2023neural,lu2023ames}, we advocate for learning the metric in a data-driven fashion via gradient descent.

Our contributions are as follows:
\vspace{-10pt}
\begin{enumerate}
\item We extend CGENNs by integrating learnable metrics, allowing the network to adapt its internal representations dynamically rather than relying on fixed, diagonal metrics.

\item  We employ eigenvalue decomposition to transform the full metric matrix representation into an intermediate computationally tractable diagonal form that can easily be integrated into CGENNs, while ensuring that the input and output data remain consistent across different geometric spaces. 

\item We leverage category theory to provide a theoretical foundation for our method. By viewing Clifford Algebras as categorical constructions, we justify the transformations applied within the network, ensuring that our approach is mathematically sound.
\end{enumerate}
\vspace{-5pt}
We validate our approach through experiments on various tasks, including $n$-body simulations, signed volume computations, and top-tagging in particle physics.

\section{Related Work}
\label{sec:Related Work}

\paragraph{The Role of Symmetries in Deep Learning} Broadly speaking, a symmetry of a system is a transformation that maintains a particular property of that system, ensuring it remains unchanged or invariant. Symmetries are widespread in real-world problems and data distributions that we aim to model in deep learning. While symmetries could be learned solely from data, integrating invariance and equivariance within artificial neural network architectures concerning specific group actions has emerged as an effective inductive bias, especially in scenarios with limited data~\citep{bronstein2021geometric}.

\paragraph{From Complex Numbers and Quaternions to Multivectors in Neural Representations} Initially, the motivation for complex-valued neural representations stemmed from their superior performance within the realm of sequence modeling and Recurrent Neural Networks (RNNs)~\citep{wisdom,unitary}. Likewise, other works also motivated them from the perspective of optimization~\citep{nitta}, generalization~\citep{hirose}, and faster learning~\citep{danihelka}. Following this line of research, complex-valued neural networks proposed a number of new building blocks that incorporated complex value representations and generalized traditional operations in neural networks such as batch normalization, weight initialization, ReLU activations, convolutions, etc., enabling them to discern intricate patterns that conventional real-valued networks might struggle to capture~\citep{trabelsi2018deep}. Quaternion based neural networks~\citep{gaudet2018deep,parcollet2019quaternion} further expanded on this line of research and introduced three imaginary components for data representation. More recently, the focus has shifted towards leveraging complex representations in the context of geometric deep learning~\citep{bronstein2021geometric} rather than in sequence modelling. Going beyond complex numbers and quaternions, Clifford algebra can encode richer representations all the way from scalars, vectors, bivectors, and trivectors, to other $k$-vectors. This capability has been showcased in studies that substitute convolution and Fourier operations in neural PDE surrogates with Clifford counterparts for both 2D and 3D tasks~\citep{brandstetter2023clifford}. Particularly relevant to our work are Clifford Group Equivariant Neural Networks (CGENNs)~\citep{ruhe2023clifford} which leverage Clifford algebra to model equivariance under orthogonal transformations of the Clifford group. However, current CGENNs are constrained by their support solely for diagonal and fixed metrics, thereby limiting the internal geometric representations they can effectively capture.

\paragraph{Metric Learning} enables models to learn how to measure similarity between data points by employing an optimal distance metric tailored to specific learning tasks, rather than relying on predefined, static metrics~\citep{Kulis2013MetricLA}. Generalizing metrics beyond simple Euclidean embeddings has been extensively studied in the literature by leveraging constant curvature Riemannian manifolds, stereographic projections, and product manifolds~\citep{gu2018learning,ganea2018hyperbolic,skopek2019mixed,de2022latent,borde2023projections,saez2024neural,kratsios2023capacity}. However, most of these works, similar to CGENN, pre-define the used metric before optimization. Other recent studies have proposed embeddings with associated differentiable metrics instead, which more closely resemble our proposed approach~\citep{lu2023ames,de2023neural}.

\paragraph{Category Theory in Deep Learning} Category theory has recently gained traction in the literature as a unifying language capable of formalizing and extending existing deep learning frameworks. For instance, \cite{gavranović2024categorical} use 2-monads to generalize geometric deep learning to non-invertible operations; \cite{fong2019backprop} formalize backpropagation as a functor; and \cite{villani2023deep} show that any deep ReLU network has a functionally equivalent three-layer network. Furthermore, category theory has also inspired novel and successful learning schemes: \cite{dehaan2020natural} use functors to construct natural graph networks, generalizing permutation equivariance; and \cite{hansen2020sheafneuralnetworks,bodnar2023neural,pmlr-v196-barbero22a,barbero2022sheaf2} augment the message-passing procedure underlying most graph neural networks with geometric information carried by sheaves, a well-known categorical construction in algebraic topology and geometry. Similar sheaf based approaches have also been extended to hypergraphs~\cite{NEURIPS2023_27f243af}.

\section{Background}
\label{sec:Background}

Next, we review the mathematical foundations of our approach, including metric spaces, Clifford algebras, eigenvalue decomposition, and category theory. We also discuss the key components comprising CGENNs, such as algebra embedding layers, generalized linear layers, geometric product layers, normalization layers, and grade-wise nonlinear activation functions.

\paragraph{Inner Product and Metric Spaces} An inner product $\langle \cdot,\cdot \rangle$ on a vector space $V$ over a field $K\in\{\mathbb{R}, \mathbb{C}\}$ is a map $V\times V\to K$ such that, for all $x,y\in V$ and $a,b\in K$, satisfies: conjugate-symmetry, $\langle x,y\rangle = \overline{\langle y,x\rangle}$ where $\overline{\,\cdot\,}$ denotes complex conjugation; linearity in the first argument, $\langle ax+by,z\rangle=a\langle x,z\rangle+b\langle y,z\rangle$; and positive-definiteness, $\langle x,x\rangle > 0$ for all $x\neq 0$. The inner product is often expressed by the matrix representation $Q$ of a bilinear form satisfying the conditions above. In this case, the inner product takes the form $\langle x,y\rangle = x^T Q y$. For example, if $Q=I$, then this is the usual dot product. A positive definite bilinear form $Q$, i.e. a matrix whose eigenvalues are all positive, induces a \textit{norm} $||\cdot||$ on $V$ by $||x||= \sqrt{x^TQx}$. As for the previous example, if $Q=I$, then this is the standard Euclidean norm. Furthermore, a norm induces a \textit{metric} (or \textit{distance}) $d$ on $V$ by setting $d(x,y)=||x-y||$, making $(V,d)$ a \textit{metric} space. However, we are mainly concerned with metrics arising from matrices $Q$, and therefore we will refer to such bilinear forms as \textit{metrics}. In this work, we focus on \textit{learning} the metric for CGNNs. During training, our algorithm initializes metrics to $Q=\text{diag}(1,\dots,1)$, the standard metric on Euclidean space, and $Q=\text{diag}(-1,\dots,1)$, the Minkowski pseudo-metric\footnote{The Minkowski metric is technically a pseudo-metric, as it violates positive-definiteness.}, and iteratively updates them via gradient-descent.

\paragraph{Clifford Algebras} A Clifford Algebra, denoted \(Cl(V, Q)\), extends the scope of classical algebra and subsumes algebraic structures such as complex numbers and quaternions. It is defined on a vector space \(V\) over a field \(K\), together with a quadratic form \(Q: V^2 \rightarrow K\) that maps pairs of vectors to the field. Its algebra operation, the geometric product, expresses geometric transformations, such as the inner $\langle.,.\rangle$ and wedge product $\wedge$ \footnote{$(v\wedge w)_{ij}=(v_iw_j-v_jw_i)$ for all $v,w\in V$. We refer the interested reader to \cite{Darling_1994}.} in algebraic terms. For all $v,w\in V$, the geometric product $vw$ is:
\begin{equation*}
    vw = \langle v, w\rangle + v\wedge w.
\end{equation*}

The Clifford Algebra $Cl(V,Q)$ is defined as a quotient of the tensor algebra $T(V)=\bigoplus_{k=0}^{\infty} T^k(V)$, where $T^k(V)=V\otimes V\dots \otimes V$ $k$ times, with the convention $T^0(V)=K$, the underlying field. The quotient defining $Cl(V,Q)$ is $T(V)$ modulo the ideal generated by $v\otimes v - Q(v,v)\cdot 1$, which essentially imposes the equation $v\otimes v = Q(v,v) \cdot 1$ on $T(V)$. Clifford Algebras also have a categorical construction (see Section~\ref{Categorical Construction of Clifford Algebras}) which will prove relevant to our methodology. 

With an orthogonal basis \(\{e_1, e_2, \ldots, e_n\}\) for \(V\), the algebra is spanned by elements of the form \(e_{i_1}e_{i_2}\cdots e_{i_k}\) where \(0 \leq k \leq n\) and \(1 \leq i_1 < i_2 < \cdots < i_k \leq n\). These span vectors, bivectors, and higher-dimensional constructs, representing directional, area, and volumetric information respectively. Hence, an element of the Clifford Algebra $Cl(V,Q)$ is called a \textit{multivector}, sometimes also referred to as Clifford numbers or multors~\citep{Snygg2011ANA}. The various dimensions (e.g., scalar, vector, bivector, etc.) of a multivector are referred to as \textit{grades}.

The dimension of a Clifford Algebra \(Cl(V, Q)\) scales as \(2^n\) for a vector space \(V\) of dimension \(n\). For instance, in a two-dimensional space (\(n=2\)), an element \(x \in Cl(V, Q)\) is represented as:

\vspace{-15pt}
\begin{equation*}
x = x^{(0)}\cdot 1 + x^{(1)}\cdot e_1 + x^{(2)}\cdot e_2 + x^{(12)}\cdot e_1e_2,
\end{equation*}
with \(x^{(i)}\in K\), \(1\) is the algebra's unit, and \(e_i\) are basis vectors of \(V\). The geometric product for multivectors distributes over $+$, analogously to the product of polynomials.
Indeed, one could compute the geometric product between multivectors $v$ and $w$ by treating each $e_i$ as a variable in a polynomial and performing polynomial multiplication without assuming commutativity of multiplication of the variables ($e_ie_j\neq e_je_i$). Additionally, the result should incorporate the relations \(v^2 = Q(v, v)\) for all vectors $v\in V$ and \(vw = -wv\) for orthogonal vectors $v,w\in V$. Hence, unlike the polynomial's case, the geometric product is not commutative.

The \textit{Clifford Group}, often denoted $Cl^{\times}(V,Q)$ is the set of elements $x\in Cl(V,Q)$ that have an inverse $x^{-1}\in Cl(V,Q)$ with respect to the geometric product, i.e. $x x^{-1}=x^{-1} x=1$. This group stands out for its capacity to algebraically represent geometric transformations such as rotations, reflections, translations and screws.

\paragraph{Norms in Clifford Algebras} While it is clear how to calculate a norm of $v\in V$ with $Q$, namely $||v||=\sqrt{v^TQv}$, we still need to specify its corresponding operation for $x\in Cl(V,Q)$. Let $\beta:Cl(V,Q)\to Cl(V,Q)$ be the \textit{main anti-involution of} $Cl(V,Q)$, the function that takes $x$ and `inverts' the order of its components: 
\begin{equation*}
    \beta(x)=\beta\left(\sum_{i\in I}c_i\,v_{i,1}\cdot v_{i,k_i}\right)=\sum_{i\in I}c_i\,v_{i,k_i}\cdot v_{i,1}.
\end{equation*}
For example, in 2 dimensions:
\begin{align*}
    \beta(x)&=\beta(x^{(0)}\cdot 1 + x^{(1)}\cdot e_1 + x^{(2)}\cdot e_2 + x^{(12)}\cdot e_1e_2)\\ 
    &= x^{(0)}\cdot e_1e_2 + x^{(1)}\cdot e_2 + x^{(2)}\cdot e_1 + x^{(12)}\cdot 1.
\end{align*}
Now, let $x_1,x_2\in Cl(V,Q)$ and $b:Cl(V,Q)\to Cl(V,Q)$ be the function: 
\begin{equation*}
    b(x_1,x_2)=\left( \beta(x_1)x_2 \right)^{(0)},
\end{equation*}
where juxtaposition is the geometric product and $(\cdot)^{(0)}$ denotes the projection on the scalar component of the resulting multivector. Finally, by denoting $x^TQx$ as $Q(x)$ for $x\in V$, we can extend this map to $\overline{Q}(x)$ for $x\in Cl(V,Q)$ by setting $\overline{Q}(x)=b(x,x)$. With this notation, the norm of $x\in Cl(V,Q)$ is $||x||=(\overline{Q}(x))^{\frac{1}{2}}$. This will be important for the normalisation operations of the CGENN.

\paragraph{Eigenvalue Decomposition} In linear algebra, a matrix \(M\) in \(K^{n \times n}\) over the field $K$ is said to be diagonalizable if there exists an invertible matrix \(P\) also in \(K^{n \times n}\) such that \(M = P^{-1}\Delta P\), where \(\Delta\) is a diagonal matrix. This process is known as the eigendecomposition of \(M\), because the diagonal entries of \(\Delta\) represent the eigenvalues of \(M\), and the columns of \(P\) correspond to their respective eigenvectors.

One can interpret \(P\) as a change of coordinates that represents a vector \(x\) in \(K^{n}\) using the coordinates defined by the eigenvectors of \(M\). Indeed, the \(i^{\text{th}}\) column of an invertible matrix \(T\) is precisely \(T e_i\), indicating where \(T\) maps the canonical basis element \(e_i\); and because a linear transformation is entirely characterised by its effect on the basis elements, this information is sufficient to fully describe the matrix's action. Thus, \(Px\) is simply \(x\) expressed in the basis provided by \(P\). Within this basis, the operation of \(M\) is expressed by the diagonal matrix \(\Delta\): \(Mx = \Delta P x = \Delta(Px)\).

Specifically, if \(K = \mathbb{R}\) and \(M\) is symmetric, the Spectral Theorem ensures that \(M\) can be diagonalized with \(P\) being orthogonal, that is, \(P^{-1} = P^T\). Now, if \(M\) functions as a metric, the inner product between two vectors \(x, y\) is calculated as follows:
\begin{equation*}
    x^T M y = x^T P^{-1} \Delta P y = x^T P^T \Delta P y = (Px)^T \Delta (Py),
\end{equation*}
indicating that when the vectors are expressed in terms of the basis provided by \(P\), the metric calculations remain consistent as if \(M\) were diagonal. This observation is crucial for our implementation since the algorithm used for metric calculations only supports diagonal metrics.

\paragraph{Basics of Category Theory}
Category theory is the study of \textit{compositionality}, where ``objects'' can be studied only in their relationships to other objects. For example, to characterise a singleton set, we will not say that it is a set $S$ with only one element, but rather, that there is a unique map from any other set to $S$.

In the first instance, we can think of a category as a collection of objects of a certain kind, such as sets, groups, and vector spaces, along with maps between them (also called morphisms or arrows) that preserve their structure, such as functions, group homomorphisms, and linear maps. Furthermore, these should be able to be composed sensibly, so that if $f:A\to B$, $g:B\to C$ and $h:C\to D$, where $A,B,C,D$ are objects in a category and $f,g,h$ are maps, then $(h\circ g) \circ f=h \circ ( g\circ f)$. A formal definition of a category is given in Appendix \ref{Background on Category Theory}.

Functors, which we will use to motivate and justify our metric learning algorithm, are maps between categories that preserve their structure. Let \(\mathcal{C}\) and \(\mathcal{D}\) be categories. A \textit{functor} between them, \(F:\mathcal{C} \to \mathcal{D}\), is an assignment of objects \(A\) in \(\mathcal{C}\) to objects \(F(A)\) in \(\mathcal{D}\), and morphisms \(f: A \to B\) in \(\mathcal{C}\) to morphisms $F(f):F(A)\to F(B)$ in $\mathcal{D}$. This assignment should also respect the structure of the categories, i.e., composition: \(F(g \circ f) = F(g) \circ F(f)\). A formal treatment of functors is given in Appendix \ref{Background on Category Theory}.

Hence, functors map one category onto another. In particular, they allow us to translate a linear map between vector spaces, a morphism in the category of vector spaces, to a map between Clifford algebras, a morphism in the category of associative algebras, thereby guaranteeing the soundness of our proposed algorithm.

\subsection{Clifford Group Equivariant Neural Networks}\label{sec:CGENN}

We now outline the main components of the CGENN, as presented in \cite{ruhe2023clifford}. The overarching idea of this neural architecture is to accept multivectors as inputs and to process them in each of their grades separately. One could visualise such a computation scheme by picturing $k$ parallel neural networks, each dedicated to one grade, interacting with each other via the geometric product layer. Importantly, each layer needs to be equivariant with respect to any Clifford group transformations.

\paragraph{Algebra Embeddings} As the CGENN accepts multivectors as inputs, we need to embed inputs $x$ into the Clifford algebra $Cl(V, Q)$. We do so with the function \(\Embed\), whose form is application-specific. For example, if $x$ is a scalar quantity (such as the charge of a particle), it is embedded as a scalar, i.e. $\Embed(x)=x\cdot 1 + 0e_1 + 0 e_2 + \dots + 0e_1e_2\dots e_n$. Alternatively, if $x$ is a point in $V$ (such as the position of a particle), it is embedded as a $1$-dimensional multivector, i.e. $\Embed(x)=0\cdot 1 + x_1e_1 +\dots +x_ne_n + 0e_1e_2 +\dots +0e_1\dots e_n$. If $x$ is a volume, it is embedded as $\Embed(x)=0\cdot 1+\dots + xe_1\dots e_n$, and so on.

\paragraph{Linear Layers} The first component of this architecture is the linear layer. For $x_1,\dots x_l\in Cl(V,Q)$, it is defined as:
\begin{equation*}
    y_{c_{\text{out}}}^{(k)} = T_{\phi_{c_{\text{out}}}}^{\text{lin}}(x_1,\dots,x_l)^{(k)}:=\sum_{c_{in}=1}^l\phi_{c_{\text{out}}c_{\text{in}}^k} x_{c_{\text{in}}}^{(k)},
\end{equation*}
where $\phi_{c_{\text{out}}c_{\text{in}}^k}$ is a learnt scalar parameter depending on the grade $k$, $c_{\text{in}}$ and $c_{\text{out}}$ are the input and output channels, and $(\cdot)^{(k)}$ is the projection on the $k^{\text{th}}$ grade. Therefore, the map $T_{\phi_{c_{\text{out}}}}^{\text{lin}}(x_1,\dots,x_l)^{(k)}$ is linear in each grade separately. This is indeed an equivariant layer, as actions of the Clifford group operate separately in each grade.

\paragraph{Geometric Product Layers}
The Geometric Product layer is the main tool that allows different grades to communicate in a geometrically principled manner. It is parameterised as follows:
\begin{equation*}
P_\phi(x_1,x_2)^{(k)}:=\sum_{i=0}^n\sum_{j=0}^n\phi_{ijk}(x_1^{(i)}x_2^{(j)})^{(k)},
\end{equation*}
where $\phi_{ijk}$ are learnable scalars for $i,j,k\in\{0,\dots,n\}$. The equivariance of this layer is guaranteed by the fact that the actions of the Clifford group respect grade projections, linearity, scalar multiplication, and products.

\paragraph{Normalisation}
To ensure numerical stability, CGENNs implement equivariant normalisation layers, defined for $x^{(m)}\in Cl^{(m)}(V,Q)$ as:
\begin{equation*}
    \text{Norm}(x^{(m)}) := \frac{x^{(m)}}{\sigma(a_m)(\overline{Q}(x^{(m)}) - 1) + 1},
\end{equation*}
where $\sigma$ is the sigmoid function and $a_m\in\mathbb{R}$ is a learnable parameter. Again, the operation is performed grade-wise and is also equivariant because $\overline{Q}(x^{(m)})$ is, as shown in \cite{ruhe2023clifford} (Theorem 3.2). Here the logistic function is used so that the denominator interpolates between $1$ (no normalisation) and $\overline{Q}(x^{(m)})$ (geometric normalisation).

\paragraph{Nonlinearities} are implemented grade-wise to maintain equivariance and are defined, for $x\in Cl(V,Q))$ as:
\begin{equation*}
    \text{NonLinear}^{(k)}(x) := \psi(f_k(x))\cdot x^{(k)},
\end{equation*}
where $\psi$ is any non-linear function $\mathbb{R}\to\mathbb{R}$ and $f_k$ is a linear function of the components of $x^{(k)}$.

\section{Method}
\label{sec:Method}

The following section details the metric learning method proposed in this work, focusing on the initialisation and processing of the metric through its eigenvalue decomposition.

\subsection{Metric Initialization}
\label{sec:metrinit}

In our method, the transition from a static metric \(Q\), typically initialized as a diagonal matrix to reflect basic geometric properties of the space (e.g., \(Q = \text{diag}(1, 1, 1)\)), to a learnable metric \(M\) involves introducing small perturbations. The process begins with \(Q\), representing the initial geometric configuration. To facilitate learning, a perturbation is added to \(Q\) through:

\begin{equation*}
    \tilde{M} = \frac{1}{2} Q + \epsilon R,
\end{equation*}

where \(R\) is a random matrix with the same dimensions as \(Q\), and \(\epsilon\) controls the amount of initial perturbation. To ensure that \(M\) is symmetric, we add the transpose of \(\tilde{M}\) to itself, since the sum of two symmetric matrices is symmetric. Hence, we obtain:

\begin{equation*}
    M = \tilde{M} + \tilde{M}^\top = Q + \epsilon (R + R^\top),
\end{equation*}

since \(Q\) is diagonal and \(Q = Q^\top\). Therefore, \(M\) is equivalent to adding ``symmetric'' noise, \(R + R^\top\), controlled by \(\epsilon\), to the initial metric. $M$ is then passed to all downstream layers. The result is a learnable metric that enables the CGENN to dynamically refine its internal geometric representation in a data-driven fashion.

\subsection{Learnable Metric via Eigenvalue Decomposition}
\label{sec:metriclearning}

The original CGENN \cite{ruhe2023clifford} implements layers consistent with a Clifford algebra $Cl(V, Q)$ for a fixed bilinear form $Q$. In particular, this bilinear form is taken as a metric, and it is used in the network's computation of \textit{norms}. This setup only supports diagonal metrics, simplifying the complexity of the space by considering distances that scale linearly along each axis independently. However, real-world data often exhibits correlations that are not captured well by such simplistic assumptions.

The transition from a diagonal to a non-diagonal metric introduces computational and theoretical challenges, particularly in the context of Clifford algebras. In this algebra, metric computations are not as straightforward as in Euclidean space. The norm calculation algorithm used in CGENNs, originally described in \cite{10.5555/1610323}, only supports diagonal metrics. 

This requires us to map a non-diagonal metric to a diagonal one in a geometrically principled way. We achieve this with the metric's eigendecomposition. A metric $M$ can be decomposed into its eigencomponents $M=P^{-1}\Delta P$, where $P$ is a matrix of eigenvectors and $\Delta$ is a diagonal matrix of eigenvalues. Therefore, we will use $\Delta$ as the diagonal matrix to compute normalisations, and carefully modify the pipeline to make it geometrically meaningful and theoretically sound. We assume that the input is given in any basis $\{e_i\}$ in $V$. The metric learning procedure can be summarised as follows:
\begin{algorithm}
\caption{Metric Learning for Clifford Group Equivariant Neural Networks -- Forward Pass}
\begin{algorithmic}[1]
\STATE $\Delta, P \gets \text{eigendecomposition}(M)$
\STATE $x \gets \Embed(x)$
\STATE $x \gets \overline{P}(x)$
\STATE $y \gets \text{CliffordNetwork}(x, \Delta, *args)$
\STATE $y \gets \overline{P^{-1}}(y)$ 
\end{algorithmic}
\label{metric-learning-algorithm}
\end{algorithm}

Next, we proceed to motivate our approach.

\paragraph{Consistency of the Input Across Spaces} One could simply adopt $\Delta$ as a diagonal metric, but in this case, we would compute norms according \textit{not} to the standard basis of $V$, but with respect to the basis given by $P$, introducing inconsistencies. Therefore, we need to express the input in the appropriate basis to make sense of $\Delta$. We do so by applying the change of coordinates $P$, as reflected in the experiments in Section \ref{Sec:experiments}, by replacing the basis vectors $e_i$ with $\xi_i = Pe_i$, which is still a basis as $P$ is invertible. We extend the linear map $x\mapsto Px$ to a map $\overline{P}:Cl(V,M)\to Cl(V,\Delta)$ such that, for $x\in Cl(V,M)$
\begin{equation*}
    x = x^{(0)}\cdot 1 + x^{(1)}\cdot e_1 + x^{(2)}\cdot e_2 + \dots + x^{(12\dots n)}e_1e_2\dots e_n,
\end{equation*}
\begin{equation*}
    \overline{P}(x) = x^{(0)}\cdot 1 + x^{(1)}\cdot \xi_1 + x^{(2)}\cdot \xi_2 + \dots + x^{(12\dots n)}\xi_1\xi_2\dots \xi_n.
\end{equation*}
This is motivated by the categorical construction of Clifford algebras (Section~\ref{Categorical Construction of Clifford Algebras}). Importantly, we preserve the algebra structure when translating multivectors from $Cl(V,M)$ to $Cl(V,\Delta)$:
\begin{gather*}
    vw=-wv \iff PvPw=-PwPv,\\
    z^2=(Pz)^2,
\end{gather*}
for any orthogonal $v,w\in V$ and any $z\in V$. The proof is given in Appendix \ref{sec:proofs}. Notably, we preserve consistency with the embedding function outlined in Section \ref{sec:CGENN}.
Operationally, when we deal with inputs $v\in V$, we can transform them with $Pv$ and embed the result, i.e.:
\begin{equation*}
    \overline{P}(\Embed(v)) = \Embed(Pv),
\end{equation*}
as $Pv=P(v_ie^i)=v_i(Pe^i)=v_i\xi^i$ using the Einstein summation convention $v_ie^i=\sum_i v_ie_i$.
Another special case is when the input $v$ is a volume. In this case, we have the following relation:
\begin{equation*}
    \overline{P}(\Embed(v)) = \Embed(\det(P)v), 
\end{equation*}
which we also prove in Appendix \ref{sec:proofs}.

\paragraph{Consistency of the Output Across Spaces} The function $\overline{P^{-1}}$ is implemented depending on the specific problem and is a crucial step of our learning procedure. It is particularly important when the output of a CGENN has a physical interpretation. Suppose that the learning task is to predict the particle's position in an $n$-body problem (Section \ref{sec:nbody}). Then, if no change of coordinates is applied, the output position is expressed in the basis $\{\xi_i\}$, but the input in the basis $\{e_i\}$. We harmonise the two by applying $\overline{P^{-1}}:Cl(V,\Delta)\to Cl(V,M)$, the extension of $P^{-1}$, which we define analogously to $\overline{P}$. For any:
\begin{equation*}
    x = x^{(0)}\cdot 1 + x^{(1)}\cdot \xi_1 + x^{(2)}\cdot \xi_2 + \dots + x^{(12\dots n)}\xi_1\xi_2\dots \xi_n,
\end{equation*}
\begin{align*}
    \overline{P^{-1}}(x) &= x^{(0)}\cdot 1 + x^{(1)}\cdot e_1 + x^{(2)}\cdot e_2 + \dots \\
    &\quad + x^{(12\dots n)}e_1e_2\dots e_n.
\end{align*}

For example, if the output of the network is: 
\begin{itemize}
    \item a \textit{point} $y\in V$ (e.g. in \ref{sec:nbody}), then $\overline{P^{-1}}$ takes the form of $y \mapsto P^{-1}y$, similarly to the classical eigendecomposition.
    \item a \textit{volume}, i.e. the top grade of the algebra (e.g. in \ref{sec:signedvolume}), then $\overline{P^{-1}}$ maps $y\mapsto \det(P^{-1})y=\frac{1}{\det(P)}y$. This is because $\det(P)$ is precisely how much an $n-$dimensional volume in an $n-$dimensional space is deformed (`stretched') by the linear transformation $P$. A proof of this is the same as for $\overline{P}$.
    \item a \textit{scalar} in $K$, then $\overline{P^{-1}}$ is the identity function, as the basis for scalars is the algebra's identity $1$.
    \item a \textit{probability} (e.g. in \ref{sec:toptagging}), then the conversion is the identity function, as in this case, the metric serves simply as a computational aid for the models' layers and there is no inherent physical meaning. 
\end{itemize}

\paragraph{Equivariance} Linear projections are generally not equivariant. This means that our change of basis via $P$ is also not equivariant, as we allow for non-diagonal metric matrices during optimization. In practice, we initialize the CGENNs to be equivariant, including the input and output transformations. We start with Euclidean or Minkowski pseudo-metrics and use these as an initial prior. However, we relax this condition and break equivariance during later stages of optimization. It is important to note that internally, the latent representations remain equivariant if we disregard the change of basis projections.

\subsection{Categorical Construction of Clifford Algebras}
\label{Categorical Construction of Clifford Algebras}

Clifford algebras have a categorical construction, which justifies and motivates the construction of Algorithm~\ref{metric-learning-algorithm}. See Appendix~\ref{Background on Category Theory} for background on category theory. A Clifford algebra $Cl(V,Q)$ is a pair $(A,i)$ with $A$ a unital associative algebra over $K$ and $i:V\to Cl(V,Q)$ a linear map with $i(v)^2=Q(v)\cdot 1_A$ for all $v\in V$ satisfying the following \textit{universal property:} given any unital associative algebra $A$ over $K$ and any linear map: 
\begin{equation*}
    j : V \to A \text{ such that } j(v)^2=Q(v)\cdot 1_A \text{ for all } v\in V,
\end{equation*} there is a \textit{unique} algebra homomorphism $f:Cl(V,Q)\to A$ such that: 
\begin{equation*}
    f\circ i = j,
\end{equation*}
i.e. $j$ factors through $Cl(V,Q)$ with $i$.

Hence, one can indeed see $Cl$ as a functor $\catname{VectQ}_K\to \catname{AssocA}$ that maps objects $(V,Q)$ to $Cl(V,Q)$ and morphisms (linear maps that preserve quadratic forms)
$f:(V,Q)\to (W,R)$ to maps:
\begin{equation*}
    \overline{f}:=Cl(f):Cl(V,Q)\to Cl(W,R).
\end{equation*}
Functoriality is guaranteed by the universal property of Clifford algebras, which ensures that maps between vector spaces preserving quadratic forms extend \textit{uniquely} to algebra homomorphisms between the respective Clifford algebras. Therefore, the maps $\overline{P}$ and $\overline{P^{-1}}$ in Algorithm \ref{metric-learning-algorithm} are precisely the extensions of the linear maps $P$ and $P^{-1}$ given by the functor $Cl$. Furthermore, our implementation of $\overline{P^{-1}}$ uses the fact that $\overline{P^{-1}}=\overline{P\,}\,^{-1}$ as per Section~\ref{sec:category-theory}

\section{Experiments}\label{Sec:experiments}

In Sections \ref{sec:signedvolume}, \ref{sec:nbody}, and \ref{sec:toptagging}, we conduct experiments with signed volumes, $n$-body problems, and top tagging, respectively. Additionally, in Section \ref{sec:Activation}, we explore the effect of activating metric learning at different stages of training and empirically examine its impact on optimization. In all experiments, we use the default configurations of the baseline CGENN \cite{ruhe2023clifford} without any hyperparameter tuning to ensure an equitable comparison.

\subsection{O(3) Experiment: Signed Volumes}
\label{sec:signedvolume}

The signed volumes experiment involves a synthetic dataset of random 3D tetrahedra. The network processes point clouds, aiming to predict covariant scalar quantities (pseudo-scalars) under $O(3)$ transformations. The prediction accuracy is measured by the mean-squared error (MSE) between the network's output and the actual signed volumes of the tetrahedra.

We compare CGENNs with a learnable metric against conventional CGENNs. Note that in this experiment, metric learning is initialised from the beginning of training alongside all other model parameters. Other baselines include a normal MLP, an MLP-based version of $E(n)$ Equivariant Graph Neural Networks~(E(n)-GNNs) \citep{satorras2021n} (this architecture leverages artificial neural networks to update positions with scalar multiplication), Vector Neurons (VNs)~\citep{deng2021vn} and Geometric Vector Perceptrons (GVPs) \citep{jing2020learning}. We train our model for $130,000$ steps, the same as the original CGENN. We calculate the mean and standard deviation with 4 different seeds.

The experimental results, as presented in Table \ref{tab:o3_results}, indicate that the learnable metric improves the performance of the original CGENN model. It also outperforms all other baselines. VNs and GVPs perform similarly with an MSE slightly lower than $10^{-1}$, while MLPs achieve better performance as the number of data samples increases, reaching MSE losses of around $10^{-3}$. Hence, all other baselines result in MSE losses that are orders of magnitude higher than those of CGENNs.

\begin{table}[h]
\centering
\caption{Test MSE loss for the synthetic $O(3)$ signed volume experiments.}
\renewcommand{\arraystretch}{1.1} 
\setlength{\tabcolsep}{1.5pt} 
\fontsize{8.5}{10}\selectfont
\begin{tabular}{lcc}
\hline
 & \multicolumn{2}{c}{MSE} \\
\cline{2-3}
Model & $n = 1000$ & $n = 65536$ \\
\hline
CGENN & $1.1\times10^{-5} \pm 1.1\times10^{-7}$ & $1.8\times10^{-6} \pm 1.1\times10^{-7}$ \\
Ours & $\mathbf{8.3\times10^{-8} \pm 1.0\times10^{-8}}$ & $\mathbf{5.1\times10^{-8} \pm 1.0\times10^{-8}}$ \\
\hline
\end{tabular}
\label{tab:o3_results}
\end{table}

\subsection{E(3) Experiment: $n$-body }
\label{sec:nbody}

The $n$-body experiment, as introduced by \citep{kipf2018neural}, sets a benchmark for assessing equivariant neural networks in the domain of physical system simulation, a topic further researched by \citep{han2022geometrically}. This experiment challenges neural architectures to predict the three-dimensional paths of~$n$ (we use $n=5$) charged particles, thereby evaluating their ability to accurately model dynamical systems. 

We compare our model against the original CGENN, lacking metric learning features, as well as steerable SE(3)-Transformers~\citep{fuchs2020se}, Tensor Field Networks~(TFNs)~\citep{thomas2018tensor}, Neural Message Passing for Quantum Chemistry Networks~(NMPs)~\citep{gilmer2017neural}, Radial Fields \citep{kohler2020equivariant}, E(n)-GNNs \citep{satorras2021n}, and Steerable $E(3)$ Equivariant Graph Neural Networks~(SEGNNs)~\citep{brandstetter2021geometric}. 

For this experiment, we train the network for $10,000$ steps over 6 different seeds to obtain the mean and standard deviation. We activate the metric learning at 90\% of the training duration. We further analyse the effect of activating metric learning earlier during the training process in Section~\ref{sec:Activation}.

The findings, as detailed in Table \ref{tab:mse_results}, indicate an improvement of our metric-augmented CGENN over the baseline CGENN and other alternative methods. It is worth noting that the CGENN presented in~\cite{ruhe2023clifford} was trained for 131,072 steps, significantly longer than ours, and nonetheless, we achieve a better performance.

\begin{table}[ht]
\centering
\caption{Test MSE loss on the \(n\)-body experiment for different methods.}
\begin{tabular}{ll}
\hline
Method & MSE ($\times 10^{-3}$) \\
\hline
SE(3)-Tr & $24.4$ \\
TFN & $24.4$ \\
NMP & $10.7$ \\
Radial Field & $10.4$ \\
E(n)-GNN & $7.0$ \\
SEGNN & $4.3$ \\
CGENN & $3.9  \pm 0.1 $ \\ 
\hline
Ours & $\textbf{3.369} \pm \textbf{0.0436}$ \\
\hline
\end{tabular}
\label{tab:mse_results}
\end{table}

\subsection{O(1, 3) Experiment: Top Tagging}
\label{sec:toptagging}

Jet tagging is a technique in collider physics for categorizing the high-energy jets spawned by particle collisions, such as those detected by the ATLAS detector at CERN~\citep{aad2008atlas, kogler2019jet}. In particular, the experiment presented here, in line with~\citep{ruhe2023clifford}, focuses on jet tagging for top quarks, the heaviest particles within the Standard Model~\citep{incandela2009status}. Our evaluation is based on the benchmark provided by~\citep{kasieczka2019top}. 

We compare CGENN with metric learning against the original CGENN model with a fixed Minkowski pseudo-metric. Due to limited computational resources, we train both models for $30\%$ of the training time reported in~\citep{ruhe2023clifford} (100,000 steps, which took 32 hours using 4 A100 GPUs and an effective batch size of 32, as opposed to the CGENN in~\citep{ruhe2023clifford}, which was trained for 331,126 steps). The metric for our model is activated $80\%$ into training. Both models are run with default hyperparameters as indicated by~\citep{ruhe2023clifford}. At least under a constrained computational budget, our method outperforms the original CGENN as shown in Table~\ref{tab:accuracy_results}, which is in line with all previous experiments as discussed in Section~\ref{sec:signedvolume} and Section~\ref{sec:nbody}.
 
\begin{table}[h]
\centering
\caption{Test accuracy on the top-tagging experiment.}
\begin{tabular}{ll}
\hline
Model & Accuracy \\
\hline
CGENN & 0.8994 \\
\hline
Ours &  \textbf{0.9102} \\
\hline
\end{tabular}
\label{tab:accuracy_results}
\end{table}

\subsection{Metric Activation}
\label{sec:Activation}

To identify the optimal timing for metric activation in CGENNs, we explore different activation regimes. Our results indicate that the timing of metric activation plays a crucial role in the efficiency of learning. The challenge in metric learning within CGENNs lies in the complexity of optimising a single metric that affects every layer of the network at once. Therefore, early activation might overcomplicate training. Conversely, late activation could function as a fine-tuning stage, making the optimisation problem more tractable.

Table \ref{tab:metric_activation} illustrates the outcomes of the $n$-body experiment, comparing the final MSE loss across three distinct metric activation timings. The results reveal that late activation, occurring at 90\% of the training duration, leads to the lowest loss.

\begin{table}[h]
\centering
\caption{$n$-body ($n=5$) test MSE across different metric activation timings.}
\begin{tabular}{ll}
\hline
Metric Activation Timing & MSE ($\times 10^{-3}$) \\
\hline
Early  (30\% of Training)       & $3.554 \pm 0.2051$           \\
Mid-Training (60\% of Training)  & $3.489 \pm 0.1190$     \\
Late  (90\% of Training)        & $ \textbf{3.369} \pm \textbf{0.0436} $  \\
\hline
\end{tabular}
\label{tab:metric_activation}
\end{table}

\subsection{Reproducibility and Hyperparameters}
\label{Reproducibility}

\( R \) is always generated from a uniform distribution with all entries sampled from the interval \([0, 1)\). Depending on the experiment, the values for \( \epsilon \) and \( Q \) are set as follows: $\epsilon = 10^{-3}$ and \( Q = \text{diag}(1,1,1) \) for the \( E(3) \) \( n \)-body and \( O(3) \) signed volume experiments, and \( \epsilon = 10^{-7} \) and \( Q = \text{diag}(1,-1,-1,-1) \)\footnote{This is equivalent to the Minkowski pseudo-metric discussed in Section~\ref{sec:Background}, but using a different sign convention.} for \( O(1,3) \) top tagging. The choice of $Q$ is in line with~\citep{ruhe2023clifford}.

\section{Conclusion \& Future Work}

Our research enhances CGENNs by integrating metric learning into the original model in a geometrically meaningful way. Although, as suggested by~\citep{ruhe2023clifford}, the CGENN formulation generalizes to orthogonal groups and preserves equivariance regardless of the metric signature, previous work fixed the metric as a predefined network configuration hyperparameter. 

In this work, instead of fixing a metric from the start, we allow the model to learn a non-diagonal metric matrix as part of the optimization process in a data-driven fashion via gradient descent. By leveraging eigenvalue decomposition to perform an internal change of basis, we use the eigenvectors to map different types of data to and from the internal neural representation back to data space, ensuring consistency of both input and output across spaces without requiring explicit modification of the original CGENN network layers, which in principle only support diagonal metrics. Additionally, we employ category theory to motivate the theoretical soundness of the approach by viewing Clifford algebras as categorical constructions. 

We validate our method empirically against different tasks, including \(n\)-body, signed volume, and top-tagging experiments. We find that enabling metric learning does indeed lead to improved performance. We also analyze the effect of making the metric learnable at different stages of the optimization process, and we empirically find that doing so towards the end of training better guarantees stable training dynamics and generally leads to better final model performance.

Future research possibilities include applying this approach to more complex datasets. Similar to \cite{ruhe2023clifford}, our evaluation did not encompass out-of-distribution tests. Finally, exploring the potential of learning a different metric at each layer may also be an interesting research avenue, but care should be taken in terms of optimization dynamics and to avoid overfitting.


\bibliography{example_paper}
\bibliographystyle{icml2024}

\newpage
\appendix
\onecolumn
\appendix

\section{Proofs for Section \ref{sec:metriclearning}}
\label{sec:proofs}
\subsection{Proof 1}
We wish to prove that, if $P$ is an orthogonal transformation, then the algebra structure is preserved, i.e.
\begin{gather*}
    vw=-wv \iff PvPw=-PwPv\\
    z^2=(Pz)^2
\end{gather*}
for any orthogonal $v,w\in V$ and any $z\in V$. Because $P:(V,M)\to(V,\Delta)$, $z^2$ is shorthand for $M(z)=z^TMz$ and $(Pz)^2$ for $\Delta(Pz)=(Pz)^T\Delta Pz$.

\textbf{Direct Proof}

$[\Rightarrow]$ Assume $vw=-wv$. We show that $PvPw=-PwPv$ by evaluating each side of the equation.
\begin{gather*}
    PvPw=\langle Pv,Pw\rangle + Pv\wedge Pw = \langle v, w \rangle + \det(P)\,v\wedge w = \det(P)\,v\wedge w\\
    -PwPv=-(\langle Pw,Pv\rangle + Pw\wedge Pv) = -\langle v, w \rangle - \det(P)\,w\wedge v = \det(P)\,v\wedge w
\end{gather*}
where we use the fact that $P$ is orthogonal, i.e. $\langle Pv,Pw\rangle=\langle v,w\rangle$, $v,w$ are orthogonal, i.e. $\langle v,w\rangle=0$, and that for any linear transformation $A$, $Av\wedge Aw=\det(A) v\wedge w$.

$[\Leftarrow]$ assume $PvPw=-PwPv$. Then, because $v,w$ are orthogonal, we get the identity 
\begin{equation*}
    \det(P)v\wedge w = -\det(P) w\wedge v\Rightarrow v\wedge w = -w\wedge v
\end{equation*}
Because $0\neq\det(P)\in\{\pm1\}$, as $P$ is orthogonal. However, with the fact that $0=\langle Pv,Pw\rangle=\langle v,w\rangle$ the LHS of the last equality is $vw$, and the RHS is $-wv$, proving the right-left direction of the `only if' part of the proposition.

Now, $z^2 = M(z) = z^TMz=(Pz)^T\Delta Pz=\Delta(z)=(Pz)^2$

\textbf{Categorical Proof}

Because $P:(V,M)\to (V,\Delta)$ is a vector space isomorphism that preserves quadratic forms, i.e. $(V,M)\cong(V,\Delta)$, and $Cl$ is a functor, we get $Cl(V,M)\cong Cl(V,\Delta)$ (Appendix \ref{sec:category-theory}) via the unique extension $\overline{P}=Cl(P)$.

\subsection{Proof 2}
We wish to prove 
\begin{equation*}
    \overline{P}(\text{Embed}(v)) = \text{Embed}(\det(P)v)  
\end{equation*}
By direct calculation:
\begin{align*}
    \overline{P}(\text{Embed}(v))= \overline{P}(v\cdot e_1e_2\dots e_n) &= v\cdot \xi_1\xi_2\dots \xi_n \\
    &= v\cdot (Pe_1)(Pe_2)\dots (Pe_n)\\
    &= \det(P)v\cdot e_1e_2\dots e_n = \text{Embed}(\det(P)v)  
\end{align*}

\section{Background on Category Theory}\label{Background on Category Theory}

Here, we provide some relevant background on Category Theory to support the discussion in Section~\ref{Categorical Construction of Clifford Algebras}.

\subsection{Basics of Category Theory} \label{sec:category-theory}
Category Theory (CT) is, essentially, the study of compositionality: the study of complex systems through their simpler parts. A key difference with set theory is that, for example, we are not allowed to inspect the internal structure of the objects. Rather, we are interested in the relationships between them.

\begin{definition}[Category]
A category $\mathcal{C}$ consists of:
\begin{itemize}
    \item a collection $ob(\mathcal{C})$ of \emph{objects}
    \item for each $A,B\in ob(\mathcal{C})$ a collection $\mathcal{C}(A,B)$ of \emph{maps} or \emph{arrows} or \emph{morphisms} from $A$ to $B$
    \item for each $A,B,C\in ob(\mathcal{C})$, a function 
        \begin{gather*}
            \mathcal{C}(A,B)\times\mathcal{C}(B,C)\rightarrow\mathcal{C}(A,C) \\
            (g,f)\mapsto g\circ f
        \end{gather*}
        called composition
    \item for each $A\in\mathcal{C}$ an element $1_A\in\mathcal{C}(A,A)$, the \emph{identity} on $A$. 
\end{itemize}
Satisfying the following axioms:
\begin{itemize}
    \item\emph{associativity}: for each $f\in\mathcal{C}(A,B)$, $g\in\mathcal{C}(B,C)$, $h\in\mathcal{C}(C,D)$: $(h\circ g)\circ f = h \circ(g\circ f)$
    \item\emph{identity laws}: for each $f\in\mathcal{C}(A,B)$, $f\circ 1_A=f=1_B\circ f$
\end{itemize}
\end{definition}

\noindent We are rather informal on the word ``collection" when referring to the objects and morphisms, as they are not necessarily sets. The reader interested in foundational issues in Mathematics can consult any Axiomatic Set Theory book, such as \cite{Takeuti1973-yj}.

\textbf{Examples of Categories} To make our definition more concrete, we present some examples categories.
\begin{enumerate}
    \item The category \catname{Set} is the category with sets as objects and functions between sets as morphisms.
    \item The category \catname{Grp} is the category with groups as objects and group homomorphisms as morphisms.
    \item The category \catname{Top} is the category with topological spaces as objects and continuous maps as morphisms.
    \item Let $k$ be a field. The category $\catname{Vect}_k$ is the category with vector spaces over $k$ as objects and linear maps as morphisms.
    \item The category \catname{1} has a single object (denoted $*$) and at least one morphisms (the identity).
\end{enumerate}

The main categories we are concerned with in this work are:
\begin{itemize}
    \item The category $\catname{VectQ}_K$ whose objects are vector spaces over a field $K$ equipped with a quadratic form and whose morphisms are linear maps preserving the quadratic forms.
    \item The category $\catname{AssocA}$ whose objects are unital associative algebras and whose morphisms are algebra homomorphisms.
\end{itemize}

\subsection{Functors}

If sets have functions relating them, categories are related by \textit{functors}. Intuitively, a functor maps objects to objects and morphisms to morphisms in a compatible way.
\begin{definition}[Functor]
    Let $\mathcal{A},\mathcal{B}$ be categories. A \emph{functor} $F:\mathcal{A}\to\mathcal{B}$ consists of:
    \begin{itemize}
        \item A function $ob(\mathcal{A})\to ob(\mathcal{B})$, written as $A\mapsto F(A)$
        \item For each $A,A'\in ob(\mathcal{A})$, a function $\mathcal{A}(A,A')\to\mathcal{B}(F(A),F(A'))$ written as $f\mapsto F(f)$ 
    \end{itemize}
    satisfying the following axioms:
    \begin{itemize}
        \item $F(f'\circ f)=F(f')\circ F(f)$ whenever $f,f'$ are composable in $\mathcal{A}$.
        \item $F(1_A)=1_{F(A)}$ whenever $A\in ob(\mathcal{A})$.
    \end{itemize}
\end{definition}

\textbf{Examples of functors} To make our definition more concrete, we present some examples of functors.
\begin{enumerate}
    \item The \emph{forgetful} functors `disregard' some properties of the objects they are mapping. For example, we can construct a forgetful functor from \catname{Grp} to \catname{Set} by mapping a group to its underlying set, and a group homomorphism to the corresponding map between sets.
    
    \item The \emph{free} functors are `adjoint' to forgetful functors. They construct objects and maps in the target category with elements that, a priori, do not have those properties. For example, one can construct a free functor from \catname{Set} to \catname{Mon}, the category of monoids: given a set $S$, its image would be the monoid consisting of formal expressions of words such as $xyz^3$ (with $x,y,z\in S$). Given a function $f:S\to S'$, its image will be the monoid homomorphism which maps words in the alphabet given by $S$ to words in the corresponding alphabet given by $f(S)$.
    \item The fundamental group $\pi$ is a functor from \catname{Top*} (pointed topological spaces) to \catname{Grp}, mapping topological spaces with a base-point to the corresponding fundamental group at that point, and a continuous map to a group homomorphism as described in any topology textbook.
    \item $T$, the tangent space, is a functor from the category of smooth manifolds with base-point \catname{Man*} to \catname{Vect}, which maps a smooth manifold with base-point to its corresponding tangent space, and a smooth map between manifolds to the corresponding tangent (linear) map: the differential. Therefore, the \textit{chain rule} for differentiable functions is just an instance of the functoriality of the tangent map.
\end{enumerate}

\begin{definition}[Isomorphism]
    A map $f:A\to B$ in a category $\mathcal{C}$ is an \emph{isomorphism} if there exists a map $g:B\to A$ in $\mathcal{C}$ such that $gf=1_A$ and $fg=1_B$. $g$ is called the \emph{inverse} of $f$ and $A$ and $B$ are said to be \emph{isomorphic}, denoted $A\cong B$.
\end{definition}

Intuitively, two objects are isomorphic if they are essentially the same, i.e. if they share the same fundamental properties in the context of the category they are in.

If $F:\mathcal{A}\to \mathcal{B}$ is a functor and $A\in ob(\mathcal{A}), B\in ob(\mathcal{B})$ with $A\cong B$, then it is easy to see that $F(A)\cong F(B)$. Further, $F(f^{-1})=F(f)^{-1}$

\end{document}